\documentclass[twocolumn, switch]{article} 

\usepackage{preprint}

\usepackage{amsmath, amsthm, amssymb, amsfonts}

\usepackage[numbers,square]{natbib}
\usepackage{natbib}

\usepackage[utf8]{inputenc}	
\usepackage[T1]{fontenc}	
\usepackage{xcolor}		
\usepackage[colorlinks = true,
            linkcolor = purple,
            urlcolor  = blue,
            citecolor = cyan,
            anchorcolor = black]{hyperref}	
\usepackage{booktabs} 		
\usepackage{nicefrac}		
\usepackage{microtype}		
\usepackage{lineno}		
\usepackage{float}			
\usepackage{amsmath}
\usepackage{multirow}
\usepackage{lipsum}		

\usepackage{newfloat}
\DeclareFloatingEnvironment[name={Supplementary Figure}]{suppfigure}
\usepackage{sidecap}
\sidecaptionvpos{figure}{c}

\usepackage{titlesec}
\titlespacing\section{0pt}{12pt plus 3pt minus 3pt}{1pt plus 1pt minus 1pt}
\titlespacing\subsection{0pt}{10pt plus 3pt minus 3pt}{1pt plus 1pt minus 1pt}
\titlespacing\subsubsection{0pt}{8pt plus 3pt minus 3pt}{1pt plus 1pt minus 1pt}

\usepackage{tikz,xcolor,hyperref}

\definecolor{lime}{HTML}{A6CE39}
\DeclareRobustCommand{\orcidicon}{
	\begin{tikzpicture}
	\draw[lime, fill=lime] (0,0)
	circle [radius=0.16]
	node[white] {{\fontfamily{qag}\selectfont \tiny ID}};
	\draw[white, fill=white] (-0.0625,0.095)
	circle [radius=0.007];
	\end{tikzpicture}
	\hspace{-2mm}
}
\foreach \x in {A, ..., Z}{\expandafter\xdef\csname orcid\x\endcsname{\noexpand\href{https://orcid.org/\csname orcidauthor\x\endcsname}
			{\noexpand\orcidicon}}
}

\title{Ditto: Motion-Space Diffusion for Controllable Realtime Talking Head Synthesis}

\author{
 Tianqi Li,\\
  Ant Group\\
  \texttt{shijian.ltq@antgroup.com} \\
   \And
 Ruobing Zheng\textsuperscript{\dag},\\
  Ant Group\\
  \texttt{zhengruobing.zrb@antgroup.com} \\
  \And
 Minghui Yang \\
  Ant Group\\
  \texttt{minghui.ymh@antgroup.com} \\
  \AND
  Jingdong Chen \\
  Ant Group \\
  \texttt{jingdongchen.cjd@antgroup.com} \\
  \And
  Ming Yang \\
  Ant Group \\
  \texttt{m.yang@antgroup.com} \\
}

\begin{document}

\twocolumn[ 
  \begin{@twocolumnfalse} 

\maketitle

\centering{\textbf{Project Page:} \href{https://digital-avatar.github.io/ai/Ditto/}{\textcolor{red}{https://digital-avatar.github.io/ai/Ditto/}}}

\hspace{\fill}

\begin{abstract}
    Recent advances in diffusion models have endowed talking head synthesis with subtle expressions and vivid head movements, but have also led to slow inference speed and insufficient control over generated results. To address these issues, we propose Ditto, a diffusion-based talking head framework that enables fine-grained controls and real-time inference. Specifically, we utilize an off-the-shelf motion extractor and devise a diffusion transformer to generate representations in a specific motion space. We optimize the model architecture and training strategy to address the issues in generating motion representations, including insufficient disentanglement between motion and identity, and large internal discrepancies within the representation. Besides, we employ diverse conditional signals while establishing a mapping between motion representation and facial semantics, enabling control over the generation process and correction of the results. Moreover, we jointly optimize the holistic framework to enable streaming processing, real-time inference, and low first-frame delay, offering functionalities crucial for interactive applications such as AI assistants. Extensive experimental results demonstrate that Ditto generates compelling talking head videos and exhibits superiority in both controllability and real-time performance.
\end{abstract}
\vspace{0.35cm}

  \end{@twocolumnfalse} 
] 


\let\thefootnote\relax\footnotetext{\dag~Corresponding Author}


\begin{figure*}
    \centering
    \includegraphics[width=1\linewidth]{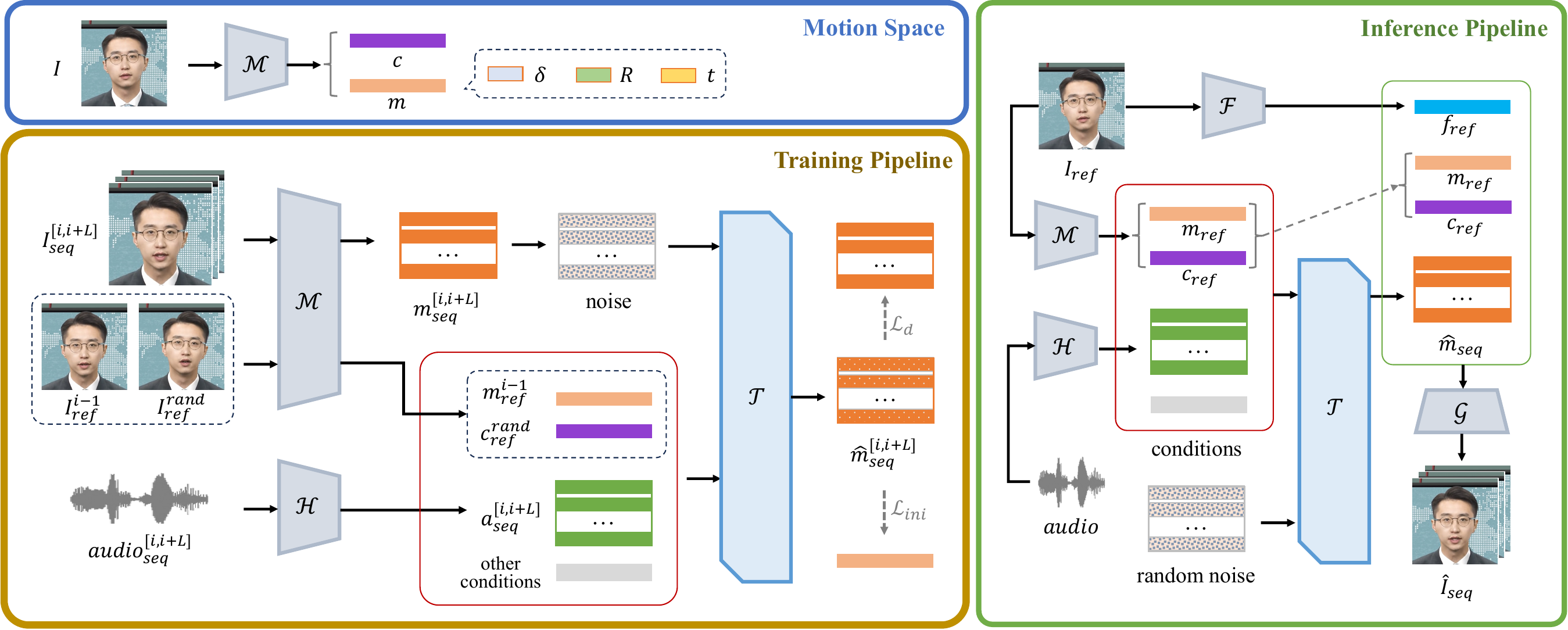}

    \caption{The summary of the proposed \textbf{Ditto}, where $\mathcal{T}$ is the DiT for motion generation, $\mathcal{H}$ is the HuBERT for audio feature extraction, $\mathcal{M}$ is the Motion Extractor, $\mathcal{F}$ is the Appearance Feature Extractor and $\mathcal{G}$ is the Face Renderer.}
    \label{fig:framework}
\end{figure*}

\section{Introduction}
Audio-driven talking head synthesis~\cite{prajwal2020lip,drobyshev2024emoportraits,tian2024emo,zhang2024learning,zheng2021learning,liu2024anitalker,corona2024vlogger} has emerged as a prominent research topic in recent years. In particular, one-shot methods~\cite{li2023one,zhang2023sadtalker,wei2024aniportrait,ye2024real3d} substantially reduce the data prerequisite for target identities, no longer relying on templates. Early approaches based on Generative Adversarial Networks (GANs)~\cite{goodfellow2014generative} are the predominant solutions, showing accurate lip synchronization but still lacking diversity and realism. Recently, diffusion-based techniques~\cite{jiang2024loopy,chen2024echomimic,xu2024hallo}, with EMO~\cite{tian2024emo} as a notable example, have made substantial progress. These methods maintain lip synchronization and capture subtle facial expressions, synchronizing head movements with speech rhythm, which are crucial traits to achieve compelling synthesized results. 

However, two critical issues hinder the broad application of these diffusion-based methods. (1) lack of control over facial motion. Existing approaches struggle to enforce fine-grained control over talking heads, including facial movements, basic emotions, and head rotations. This leaves users with no direct means to adjust the results in addition to regeneration. As the generation quality is relatively random, achieving desired results becomes challenging and potentially time-consuming. (2) Slow inference speed. Currently, most methods can hardly achieve real-time inference on a single GPU, which is a critical requirement for interactive scenarios, such as AI assistants and live video streaming. VASA-1~\cite{xu2024vasa} innovatively achieves real-time inference through a two-stage generation approach, demonstrating the feasibility of training a Diffusion Transformer (DiT)~\cite{peebles2023scalable} based on a motion space. However, due to its source code is not publicly available, related follow-up work is limited. On the other hand, VASA-1 uses an implicit motion representation, which does not support control or adjustment of the generated results.

In this paper, we propose Ditto, a diffusion-based talking head framework that enables fine-grained controls and real-time inference. Specifically, we utilize an off-the-shelf motion extractor~\cite{guo2024liveportrait} and devise a conditional diffusion transformer to generate representations in a specific motion space. We optimize the model architecture and training strategy to address inherent issues in motion representations, including insufficient disentanglement between motion and identity information, and large internal discrepancies within the motion representations.

Besides, we employ diverse conditional signals to achieve fine-grained control over the generation process. We also establish a mapping between motion representation and facial semantics, enabling fine-tuning of facial movements and correction of visual defects. 

Moreover, we jointly optimize the holistic framework to enable streaming processing, real-time inference, and low first-frame delay, offering functionalities crucial for interactive applications such as AI assistants.  Extensive experimental results demonstrate that Ditto generates compelling talking head videos and exhibits superiority in both controllability and real-time performance. For the advancement of the community, we make our source code available for open-source use.

\section{Related Work}
In audio-driven talking head synthesis, one-shot methods have garnered research attention due to their ability to minimize data requirements to a single photograph. While early GAN-based approaches have succeeded in generating photorealistic textures and accurate lip movements, they fall short in capturing realistic expressions and head movements. Recent diffusion-based methods have made substantial progress in addressing these limitations, truly achieving the generation of realistic talking-head videos without relying on carefully curated templates. EMO~\cite{tian2024emo} stands out as a groundbreaking approach that leverages audio and other ``weak controls'' to train a conditional diffusion model for end-to-end video generation, enhancing the vividness and realism of the generated results. Subsequent methods have widely adopted its innovative Dual U-Net architecture and two-stage training scheme. Notable follow-up works include EchoMimic~\cite{chen2024echomimic}, Hallo~\cite{xu2024hallo}, and Loopy~\cite{jiang2024loopy}, which have improved the original framework in various aspects such as hierarchical audio-visual mapping, diverse driving signals, and long-term motion modeling, respectively. Nevertheless, these methods still rely on denoising and rendering images in a general VAE space. This redundant, implicit latent space not only increases the learning complexity of the Diffusion model but also results in slow inference speeds. VASA-1~\cite{xu2024vasa} introduces a novel approach by training a Diffusion Transformer (DiT)~\cite{peebles2023scalable} within a motion-appearance disentangled space, defined by a face reenactment model~\cite{drobyshev2022megaportraits}. This latent space, tailored specifically for the facial domain, combined with a two-stage generation process, significantly reduces inference time while maintaining vibrant results.

\section{Method}

In this section, we introduce the detailed implementation of the Ditto framework. We begin with a brief introduction to the motion space. Then we discuss the model architecture and training strategy to address inherent issues in motion representations. Finally, we introduce a novel motion control approach and demonstrate how to correct visual defects using gaze as an example.

\subsection{Motion Space}
While Latent Diffusion Models (LDMs)~\cite{Rombach_Blattmann_Lorenz_Esser_Ommer_2022} significantly reduce computational requirements by operating in latent space rather than pixel space, their pretrained VAE-defined latent representations present two key limitations in talking head synthesis. (1) The latent space remains redundant, with entangled motion and texture, introducing overhead in both diffusion training and pixel-level rendering. (2) These representations are implicit and lack explicit correspondence with facial attributes.

Due to the shared underlying facial geometry, individuals across different identities exhibit highly consistent facial motion patterns. Recent facial reenactment works~\cite{Wang_Mallya_Liu_2021,hong2022depth,siarohin2019first,siarohin2021motion,zhao2022thin,zhang2023metaportrait} also fully utilize this characteristic. We construct the motion space based on LivePortrait~\cite{guo2024liveportrait}, one of the state-of-the-art methods in this field. We aim to let the diffusion model generate universal motion, and then incorporate identity information during pixel-level rendering. 

Specifically, a single frame $I$ is processed through Motion Extractor $\mathcal{M}$ derived from~\cite{guo2024liveportrait} to yield canonical keypoints $\mathbf{c}\in \mathbb{R}^{K \times 3}$ , expression deformations $\boldsymbol{\delta}\in \mathbb{R}^{K \times 3}$, and headposes $\mathbf{R}\in \mathbb{R}^{3 \times 3}$ with translations $\mathbf{t}\in \mathbb{R}^{3}$. The $\mathbf{m} = \{\boldsymbol{\delta}, \mathbf{R}, \mathbf{t} \}$ serve as the identity-agnostic motion representation, which is used for training the diffusion model to predict the corresponding audio-driven $\mathbf{\hat{m}} = \{\boldsymbol{\hat{\delta}}, \mathbf{\hat{R}}, \mathbf{\hat{t}} \}$, as illustrated in Figure~\ref{fig:framework}.

Then a one-shot face renderer, comprising an Appearance Feature Extractor $\mathcal{F}$ and a Face Renderer $\mathcal{G}$~\cite{guo2024liveportrait}, is used to synthesize talking head videos. Given a target portrait, $\mathcal{M}$ first extracts its canonical keypoints $\mathbf{c_{ref}}$ and motion representation $\mathbf{m_{ref}} =\{\mathbf{\boldsymbol{\delta}_{ref}}, \mathbf{R_{ref}}, \mathbf{t_{ref}} \}$. Then the reference and generated implicit 3D keypoints ($\mathbf{x_{ref}}$ and $\mathbf{\hat{x}}$) are computed as:
\begin{align}
\mathbf{x_{ref}} &= \mathbf{c_{ref}} \mathbf{R_{ref}} + \mathbf{\boldsymbol{\delta}_{ref}} + \mathbf{t_{ref}}, \\
\mathbf{\hat{x}} &= \mathbf{c_{ref}} \mathbf{\hat{R}} + \boldsymbol{\hat{\delta}} + \mathbf{\hat{t}}.
\end{align}

These two sets of implicit keypoints guide $\mathcal{G}$ in learning a warping field that transforms the appearance features $\mathbf{f_{ref}}$ from $\mathcal{F}$ to fit the generated motions and decodes the warped features into image space to synthesize frame $\hat{I}$ as:
\begin{align}
\hat{I} &= \mathcal{G} \left( \mathbf{f_{ref}}, \mathbf{x_{ref}}, \mathbf{\hat{x}}  \right).
\end{align}

\subsection{Motion Generation with Diffusion Transformer}

\begin{figure}
  \centering
  \includegraphics[width=1\linewidth]{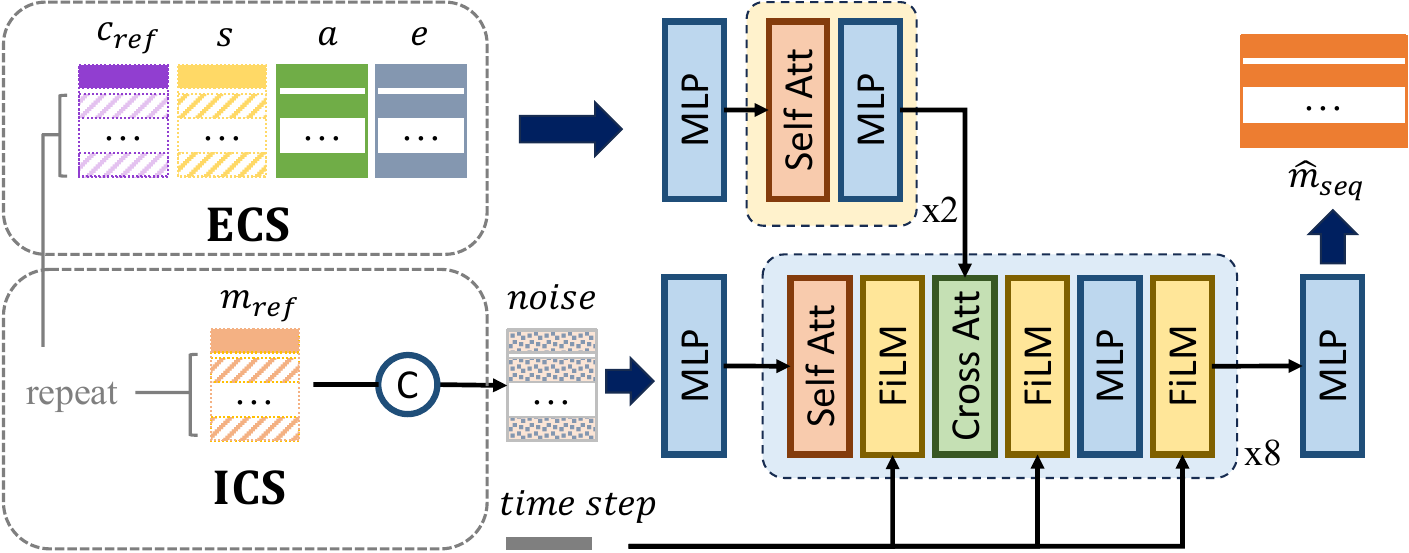}

  \caption{Visualization of the architecture of the proposed conditional Diffusion Transformer, which generates compact motion representations based on various conditional signals.}
  \label{fig:dit}
\end{figure}

As shown in Figure~\ref{fig:dit}, we employ a Conditional Diffusion Transformer (DiT) for audio-to-motion generation. Along with the \textbf{audio feature} $\mathbf{a}$, we incorporate multiple auxiliary conditional signals as follows.

\subsubsection{Conditional Signals}

Considering the fact that even LivePortrait~\cite{guo2024liveportrait} is hard to fully disentangle motion and identity information, and the impure motion representations increase the difficulty of the training process. We use the canonical keypoints as the \textbf{identity feature} $\mathbf{c_{ref}}$ to guide the generation of facial motions, ensuring the generated motions match the target facial characteristics and can appropriately animate the target identity. This identity-adaptive setting mitigates the long-standing issue of incomplete disentanglement between motion and identity~\cite{deng2024portrait4d,tran2024voodoo}. Ablation experiments also confirm the above conclusions.

We also extract emotions~\cite{savchenko2022hsemotion} from video frames and employ a unique \textbf{emotion label} $\mathbf{s}$ at the clip level, establishing a direct and controllable correspondence instead of simply learning the audio-to-expression mapping. \textbf{Eye state} $\mathbf{e}$ is composed of the aspect ratio and relative pupil position~\cite{lugaresi2019mediapipe}, governing blinking and gaze which have weak correlations with audio. A \textbf{reference initial motion} $\mathbf{m_{ref}}$ serves as a guide for the motion generation within each clip, which can be utilized to enhance inter-clip motion continuity, as well as to periodically steer the motion towards a specified target state, thereby mitigating error accumulation in long sequence generation~\cite{wang2024unianimate}. 

For $\mathbf{e}$, $\mathbf{c_{ref}}$, and $\mathbf{s}$, we use them as the Enhanced Conditional Signals (ECS) which are aligned temporally with audio features $\mathbf{a}$, followed by channel-wise concatenation before feeding into a cross-attention module to guide the generation process. As for $\mathbf{m_{ref}}$, we use it as an Initial Conditional Signal (ICS) by replicating and concatenating it channel-wise with the noise sequence, aiming to influence motion generation primarily in the initial stages. In summary, we denote our input condition as $\mathbf{C} = \left\{\mathbf{a},\mathbf{e}, \mathbf{c_{ref}},\mathbf{s},\mathbf{m_{ref}} \right\}$

\subsubsection{Training  Strategy}
We devise specialized strategies for training the DiT to accommodate the specific motion representations.

\textbf{Horizontal Flip}: Facial motions associated with speech are predominantly symmetrical. However, in-the-wild talking-head training data often exhibits an uneven distribution of head orientations. This imbalance can affect the extracted motion features, potentially biasing generated motions towards one side. To address this issue, we horizontally flipped the face images, to augment the training data to a balanced audio-to-motion correspondence for both sides. 

\textbf{Adaptive Loss Weights}: Different components of the motion representation exhibit variations in their movement patterns. For example, motions such as lip movements, facial expressions, eye blinking, and head pose differ significantly in their relationships with driving signals like audio, and in the magnitude of their movements. It is difficult to train the DiT to weigh all motion representations uniformly. Therefore, we group the motion representations based on their control regions, and dynamically adjust the weights of each group throughout the training process. We calculate the difference in average loss between the current epoch and the previous epoch, then apply softmax normalization across groups to derive adjustment coefficients for the loss weights. Experiments demonstrate that this dynamic weighting strategy improves both the loss convergence and the quality of generated motion.

\begin{figure}[t]
  \centering
  \includegraphics[width=1\linewidth]{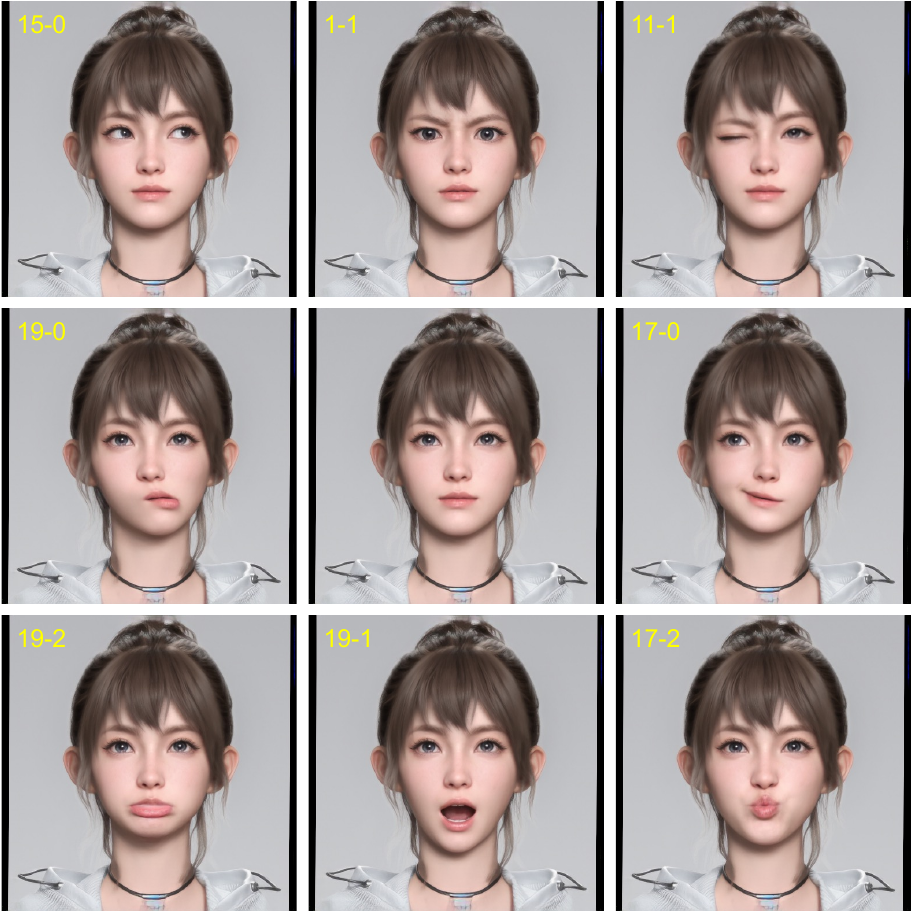}

  \caption{Examples of dedicated fine-grained control of subtle facial motion. The notation ``15-0'' in the top-left image denotes the application of an x-axis offset to the 15th keypoint. The central image depicts the rendered result when applying zero offsets across all dimensions.}
  \label{fig:bs}
\end{figure}

\textbf{Validation Metric}: Given that lip movements are the primary descriptor of facial motion, we employ a lipsync score~\cite{prajwal2020lip} as our validation metric. We render the generated motions from the validation videos and use this metric to assist in selecting the preferred model checkpoint. This approach addresses a common training issue in diffusion models, where the loss curve often fails to reflect the generation quality accurately.

\textbf{Loss Function}: 
Following DDPM~\cite{Ho_Jain_Abbeel_Berkeley}, a denoising network $\mathcal{T}$ is trained by minimizing the mean square error loss as follows:
\begin{align}
  \mathcal{L}_d = \mathbb{E}_{t \sim  \mathcal{U}[1, T], \mathbf{m}_0, \mathbf{C}} \left[ \left\| \mathbf{m}_0 - \mathcal{T}(\mathbf{m}_t, \mathbf{C}, t) \right\|_2^2 \right],
\end{align}
where $\mathbf{C}$ is the conditional signal,  $\mathbf{m}_t$ is a perturbed version of real motion data $\mathbf{m}_0$ by adding $t$ step noises. To enhance temporal stability, we also regress the velocities and accelerations of motion movements.
\begin{align}
\mathcal{L}_t = \left\| \mathbf{\hat{m}^{\prime}} - \mathbf{m}^{\prime}  \right\|_2^2 + \left\| \mathbf{\hat{m}^{\prime\prime}} - \mathbf{m}^{\prime\prime}  \right\|_2^2  ,
\end{align}
where $\mathbf{m}^{\prime\prime}$ and $\mathbf{m}^{\prime}$ denote the first-order and second-order derivatives of motion representations $\mathbf{m}$. Furthermore, to enhance the initial motion guidance, we compute an initial motion loss $\mathcal{L}_{ini}$ between the reference initial motion and the first element of the generated motion clip. In summary, the final loss can be expressed as:
\begin{align}
\mathcal{L} = \mathcal{L}_{d} + \mathcal{L}_{t} + \mathcal{L}_{ini} 
\end{align}

\subsubsection{Motion Control}

Unlike headpose which can be directly controlled via rotation and translation, our implicit keypoint-based motion representations~\cite{Wang_Mallya_Liu_2021,hong2022depth,siarohin2019first} lack clear correspondence with facial attributes. To address this limitation,~\cite{guo2024liveportrait} employs landmark-guided optimization to align implicit keypoints with explicit landmarks. We extend this work by establishing the direct mapping between deformations and facial semantics, enabling motion control in a blendshape-like manner. Specifically, we represent expression deformation as a 63-D vector, corresponding to the x, y, and z coordinates of 21 implicit 3D keypoints. For each dimension, we apply small positive and negative offsets and render facial images for a fixed identity. As illustrated in Figure~\ref{fig:bs}, each dimension of deformations affects relatively distinct facial regions. For instance, the 34th dimension governs the right eye's opening and closing motion. The 58th dimension controls mouth open, analogous to the ``jaw open'' blendshape in ARKit. Therefore, we enable two types of generation control: (1) Regional control, where motion generation can be constrained to specific local facial areas. For example, limiting movements to facial features close to the neck for better integration with the torso, and (2) Magnitude control, where we can impose constraints on deformation values to prevent artifacts such as unnatural facial expressions.

Next, we introduce how to correct visual defects in the generated motions through a case of gaze adjustment. In the early test, we discover that the generated gaze is bound to the head pose. This results in the avatar's gaze constantly wandering with head movements in the generated video, unable to focus on the camera. The reason is that we control the gaze by adding the per frame eye state $\mathbf{e}$  during training. But we use the fixed eye state during inference, which is extracted from the single reference image. 

To address this issue, we record a template video, requiring the actor to smoothly rotate head in various angles while maintaining eye focus on the camera and a constant facial expression. Then we extract the corresponding head poses $\mathbf{R}$ and expression deformations $\boldsymbol{\delta}$. By subtracting the initial frame's $\mathbf{R_0}$ and $\boldsymbol{\delta_0}$ from the $\mathbf{R}$ and $\boldsymbol{\delta}$ of each frame, we obtain synchronized variation data for pose and motion. Since the actor's expression remains constant throughout the video, the motion variation only reflects the change in gaze. We use a regression $\boldsymbol{\delta_e} = \mathcal{K}(\boldsymbol{R_e})$ to obtain the mapping between the gaze variation $\boldsymbol{\delta_e}$ and the head pose variation $\boldsymbol{R_e}$, which is used to correct the generated gaze based on the current headpose, $\boldsymbol{\delta}_{correct} = \boldsymbol{\delta} + \boldsymbol{\delta_e}$.

\begin{table*}
  \caption{\textbf{Comparisons with existing methods on the Talk9 dataset}. The s* represents the number of denoising steps. The \textbf{best} and \underline{second best} results are in \textbf{bold} and \underline{underline} specifically.}
  \label{tab:comp}
  \setlength{\tabcolsep}{4.8mm}
  \centering
  \begin{tabular*}{0.85\linewidth}{lccccc|c}
  \toprule
  Method        &  FID $\downarrow$  &  FVD $\downarrow$  &  CSIM $\uparrow$  &  Sync-C $\uparrow$  &  Sync-D $\downarrow$ & RTF $\downarrow$ \\
  \midrule
  GT        &  -      &  -      &  -    & 8.044 & 6.943 & - \\
  \midrule
  MuseTalk  & 21.445 	& 436.862 &	0.807 &	5.586 &	8.400 & 2.248 \\
  EchoMimic &	42.554 	& 395.754 &	0.840 &	5.733 &	9.204 & 35.528 \\
  Hallo     &	22.996 	& 271.680 &	0.812 &	7.652 &	7.590 & 53.082 \\
  Hallo2	  & 22.899 	& 245.236 &	0.806 &	7.737 &	7.608 & 56.838 \\
  \midrule
  Ours-s50      &	\underline{17.254} 	& \textbf{219.368} &	\textbf{0.864} &	\underline{8.069} &	\textbf{7.114} & \underline{2.121} \\
  Ours-s10	  & \textbf{17.060} 	& \underline{231.182} &	\underline{0.861} &	\textbf{8.111} &	\underline{7.291} & \textbf{0.635} \\
 
  \bottomrule
\end{tabular*}
\end{table*}

\begin{table}
  \caption{The quantitative comparisons on the HDTF100 dataset. \textit{Data with $\dag$ are from Hallo2 paper.}}
  \label{tab:comp-hdtf}
  \centering
  \begin{tabular}{lccccc}
  \toprule
  Method        &  FID $\downarrow$  &  FVD $\downarrow$  &  SyncC $\uparrow$  &  SyncD $\downarrow$ \\
  \midrule
  EchoMimic $^\dag$  & 47.331 &	532.733 &	5.930 &	9.143  \\
  Hallo $^\dag$  &	16.748 &	366.066 &	7.268  &	7.714  \\
  Hallo2 $^\dag$	& \underline{16.616} &	\underline{239.517} &	\underline{7.379}  &	\underline{7.697}  \\
  Ours &	\textbf{16.430} &	\textbf{134.640} &	\textbf{8.939}  &	\textbf{6.913}  \\
  \bottomrule
\end{tabular}
\end{table}

\subsection{Realtime Streaming Inference}

Realtime dialogue scenarios impose two critical requirements on talking head methods: (1) low-latency inference within each module and (2) efficient data streaming between modules. To cope with these requirements, we optimize the inference process across three primary modules: audio feature extraction, motion generation, and video synthesis.

\subsubsection{Audio Feature}
We employ the HuBERT~\cite{hsu2021hubert} for audio feature extraction, incorporating transformer~\cite{vaswani2017attention} acceleration techniques to enable realtime streaming audio processing. Specifically, we adopt a KV cache strategy~\cite{dai2019transformer} to provide pseudo-contextual information extracted from a fixed corpus, concatenated with incoming audio stream segments. This approach allows the HuBERT model pre-trained on long utterances to maintain high-quality features even on extremely short audio segments. Complementarily, we implement causal mask to ensure proper attention allocation to the actual input while significantly reducing computational complexity. These optimizations enable our model to process 0.4-second audio stream units in realtime on CPU environment. We utilize these enhanced features for both model training and inference.

\subsubsection{Motion Generation}
We align the audio features with the video frame rate and segment the streaming audio features into fixed-length intervals, incorporating a predefined overlap between adjacent segments. We leverage the long video generation strategy introduced in~\cite{zhang2024mimicmotion}, where we substitute progressive latent fusion with segment-wise fusion on generated motion sequences to support streaming output. During fusion, weights are determined by the relative position of each frame within its segment, with frames closer to the segment center receiving higher weights.

To achieve realtime performance, we reduce the number of denoising steps in DiT inference from 50 to 10. Benefiting from the motion representation, the inference results at 10 and 50 steps maintain comparable quality in evaluation. The DiT model is converted to TensorRT for execution on GPU environment.

\subsubsection{Video Synthesis}
Our one-shot face renderer sequentially generates photorealistic images from motion sequences. Features relevant to the target identity are pre-extracted and stored. The TensorRT-optimized renderer enables realtime inference and video stream output on GPUs. If video output is required, a CPU-based FFmpeg can parallelly compress the video stream.

\section{Experiments}
\subsection{Implementation Details}
For training data, we collect and clean about 50 hours of broadcast scene videos from 330 identities, with an average video length of 150 seconds. 
For evaluation, due to the limited computing resources, we conduct our primary quantitative comparison on \textit{Talk9} dataset while providing additional quantitative comparison on \textit{HDTF100} dataset. The \textit{Talk9} dataset consists of 9 videos from~\cite{guo2021ad, lu2021lsp, shen2022learning, li2023efficient} and has been widely used in GAN/NeRF-based talking head evaluations. \textit{HDTF100} contains 100 randomly sampled videos from the HDTF~\cite{zhang2021flow} dataset, which has been widely used for quantitative evaluation in previous works~\cite{chen2024echomimic, xu2024hallo, cui2024hallo2}.
To evaluate the generalizability of our method, we select diverse image and audio samples from~\cite{xu2024hallo,chen2024echomimic} for qualitative evaluations 

In experiments, the headpose euler angles are represented using discrete bins~\cite{Wang_Mallya_Liu_2021}, and the final motion representation contains 265 dimensions. We have trained our models using 8 NVIDIA A100 GPUs with a batch size of 1024, using the Adan~\cite{xie2024adan} optimizer with a learning rate of 1e-4 and 0.02 weight decay for 500 epochs. We select the preferred checkpoint based on the lipsync score. 

During training, audio features, video frames, and corresponding conditional signals are aligned at 25 fps. We sample segments of $L$=80-frames (3.2s) from video clips to train the Diffusion Transformer. The reference initial motion $\mathbf{m_{ref}}$ is derived from the preceding frame of the current training segment, and the reference canonical keypoints $\mathbf{c_{ref}}$ are taken from a random frame of the same video clip. During inference, both the $\mathbf{m_{ref}}$ and $\mathbf{c_{ref}}$ are derived from the source image.

\subsection{Baseline and Metrics}
We compare our method with recent publicly available implementations, including the non-diffusion-based lip editing method MuseTalk~\cite{zhang2024musetalk}, and the diffusion-based talking head generation methods EchoMimic~\cite{chen2024echomimic}, Hallo~\cite{xu2024hallo} and Hallo2~\cite{cui2024hallo2}.
The following metrics are used to evaluate the generated results: Fréchet Inception Distance (FID)~\cite{heusel2017gans} for frame-level quality, Fréchet Video Distance (FVD)~\cite{unterthiner2019fvd} for video-level quality, Cosine Similarity (CSIM) of face recognition features~\cite{deng2019arcface} for identity consistency, Sync-C and Sync-D~\cite{prajwal2020lip} for audio-visual synchronization. 
We also evaluate the inference performance by Real-Time Factor (RTF) and First-Frame Delay (FFD). $\text{RTF}<1$ and an FFD as low as possible are the basic requirements for realtime streaming inference.

\begin{table}
  \caption{User study results.}
  \label{tab:comp-mos}
  \centering
  \begin{tabular}{lcccc}
  \toprule
  Method        &  Visual quality  &  Lipsync  &  Naturalness \\
  \midrule
  EchoMimic & 54.0\% &	38.0\% &	58.7\%  \\
  Hallo  &	21.3\% &	36.0\% &	33.3\%   \\
  Hallo2 & 40.7\%  &	45.3\%  &	\textbf{59.3\%}  \\
  Ours &	\textbf{84.0\%} &	\textbf{80.7\%}  &	48.7\%  \\
  \bottomrule
\end{tabular}
\end{table}

\begin{figure*}
  \centering
  \includegraphics[width=0.95\linewidth]{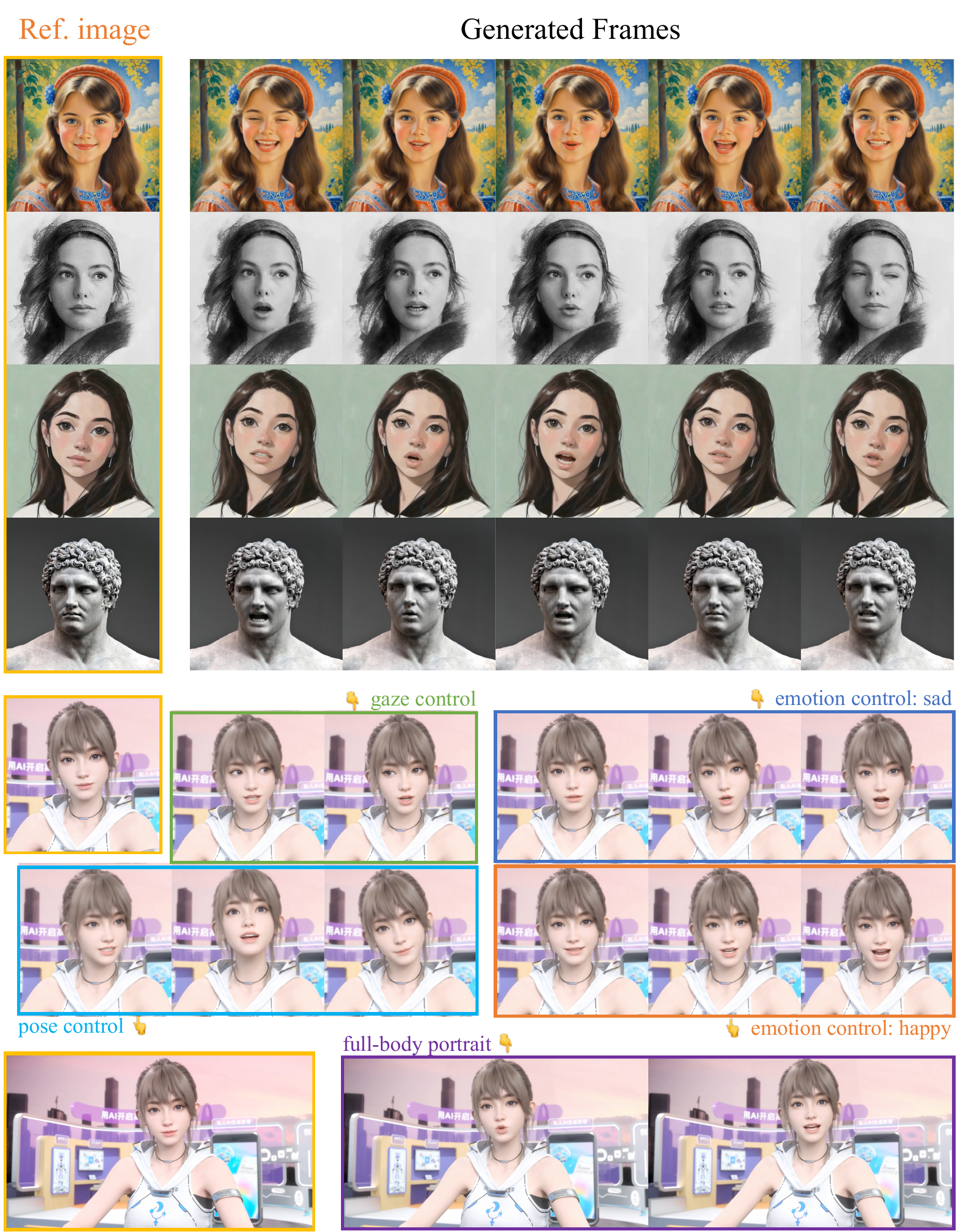}
  \setlength{\abovecaptionskip}{2mm}
  \caption{Generation results with portraits of different styles and scales. Fine-grained control over gaze, emotion, pose, etc.}
  \label{fig:res}
\end{figure*}

\begin{figure*}
  \centering
  \includegraphics[width=1\linewidth]{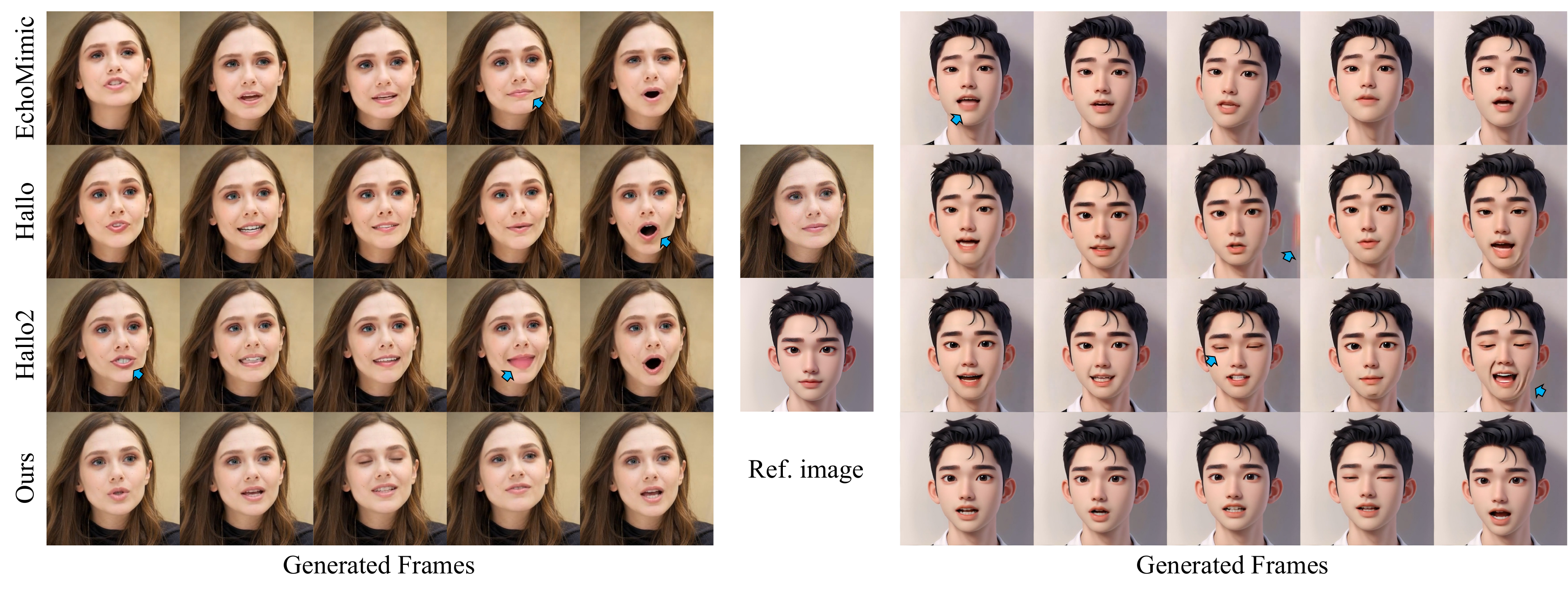}

  \caption{The qualitative comparison of our approach on two characters with different styles, poses, and hairstyles. For each character, we generate videos using the same audio input with each method and select frames at the same location for comparison. Blue arrows indicate the locations of artifacts, including inaccurate lip movements, disordered teeth, blur, etc.}
  \label{fig:comp}
\end{figure*}

\subsection{Quantitative Evaluation}

The quantitative results on Talk9 and HDTF100 datasets are shown in Table~\ref{tab:comp} and Table~\ref{tab:comp-hdtf}, respectively.
Our method achieves the best results among competing methods on all evaluated metrics. The motion representation facilitates the learning of the diffusion model and makes the lipsync score close to the real video, which is the most important criterion for evaluating talking head tasks. How to maintain identity consistency, measured by the CSIM score, is a long-standing challenge for one-shot methods. Our method achieves higher CSIM scores than other methods, attributed to the disentanglement of appearance and motion. Our method also effectively improves the frame quality and inter-frame continuity of the generated video obtaining the lowest FID and FVD among all methods. Table~\ref{tab:comp} also reports the results of reducing the number of denoising steps. The results of 10-step and 50-step denoising exhibit minimal disparity in end-to-end evaluation, substantiating the efficacy of our inference optimization. Analysis of the RTF reveals that diffusion-based methods exhibit inference speeds 30-50 times slower than realtime. In contrast, our proposed method achieves inference capabilities that surpass realtime performance, and even outpace the non-diffusion-based lip-editing method MuseTalk.

\paragraph{User Study.}
To estimate the quality of our method and SOTAs from human perspectives, we conduct a blind user study with 10 participants. Specifically, we randomly select 5 image-audio pairs to generate 5 driving clips for each tested method, resulting in a total of 20 clips. Each participant is presented with two videos generated by different methods for the same set of inputs, and asked to choose the better one in terms of visual quality, lip synchronization, and naturalness of facial expressions and head movements. This process is repeated $C_4^2$ times. The results are summarized in Table~\ref{tab:comp-mos}, where our method outperforms other
methods in visual quality and lip synchronization. We also notice that Ditto performs slightly weaker in the measure of ``naturalness of facial expressions and head movement''. We believe this is due to the motion representation attenuating the learning of some high-frequency motion information, which still requires our continuous optimization.

\begin{table}
  \caption{Ablation study results. \textbf{C-kp} represents canonical keypoints condition, \textbf{Emo} represents emotion label condition, and \textbf{Ada-w} represents Adaptive Loss Weights.}
  \label{tab:abs}
  \centering
  \setlength{\tabcolsep}{1.5mm}
  \begin{tabular*}{1\linewidth}{lccccc}
  \toprule
  Method        &  FID $\downarrow$  &  FVD $\downarrow$  &  CSIM $\uparrow$  &  SyncC $\uparrow$  &  SyncD $\downarrow$ \\
  \midrule
  Full Model & 17.254 &	219.368 &	0.864 &	8.069 &	7.114  \\
  w/o C-kp  &	17.438 &	261.693 &	0.849 &	6.932 &	8.476  \\
  w/o Emo	& 23.036 &	324.078 &	0.792 &	6.310 &	8.990  \\
  w/o Ada-w &	18.978 &	588.821 &	0.844 &	0.289 &	14.992  \\
  \bottomrule
\end{tabular*}
\end{table}

\subsection{Qualitative Evaluation}

The qualitative results are presented in Figures~\ref{fig:res} and Figure~\ref{fig:comp}.
Figure~\ref{fig:comp} shows a qualitative comparison of our approach on two characters with different styles, poses, and hairstyles. For each character, we generate videos using the same audio input with each method and select frames at the same location for comparison. Our method shows advantages in the consistency of generated details. For example, the shape and texture of teeth generated by baselines change between frames.
The baselines also have problems such as generating exaggerated expressions, blur, and inaccurate lip movements. In addition, the generation results of Hallo and Hallo2 for stylized portraits may deteriorate gradually over time, such as excessive facial wrinkles, abnormal shadows, and background artifacts. In contrast, our method shows stable performance across diverse portrait styles.

Figure~\ref{fig:res} demonstrates that Ditto can not only handle portraits of different styles but also support fine-grained generation control, such as gaze, pose, emotion, etc.
Furthermore, our approach enables seamless stitching of the generated head with the original body, facilitating full-body portrait animation without the misalignment issues often associated with end-to-end diffusion methods. This makes Ditto readily integrate with the body-driven model, achieving strong generation capabilities in a wide range of applications.

\begin{figure}
  \centering
  \includegraphics[width=1\linewidth]{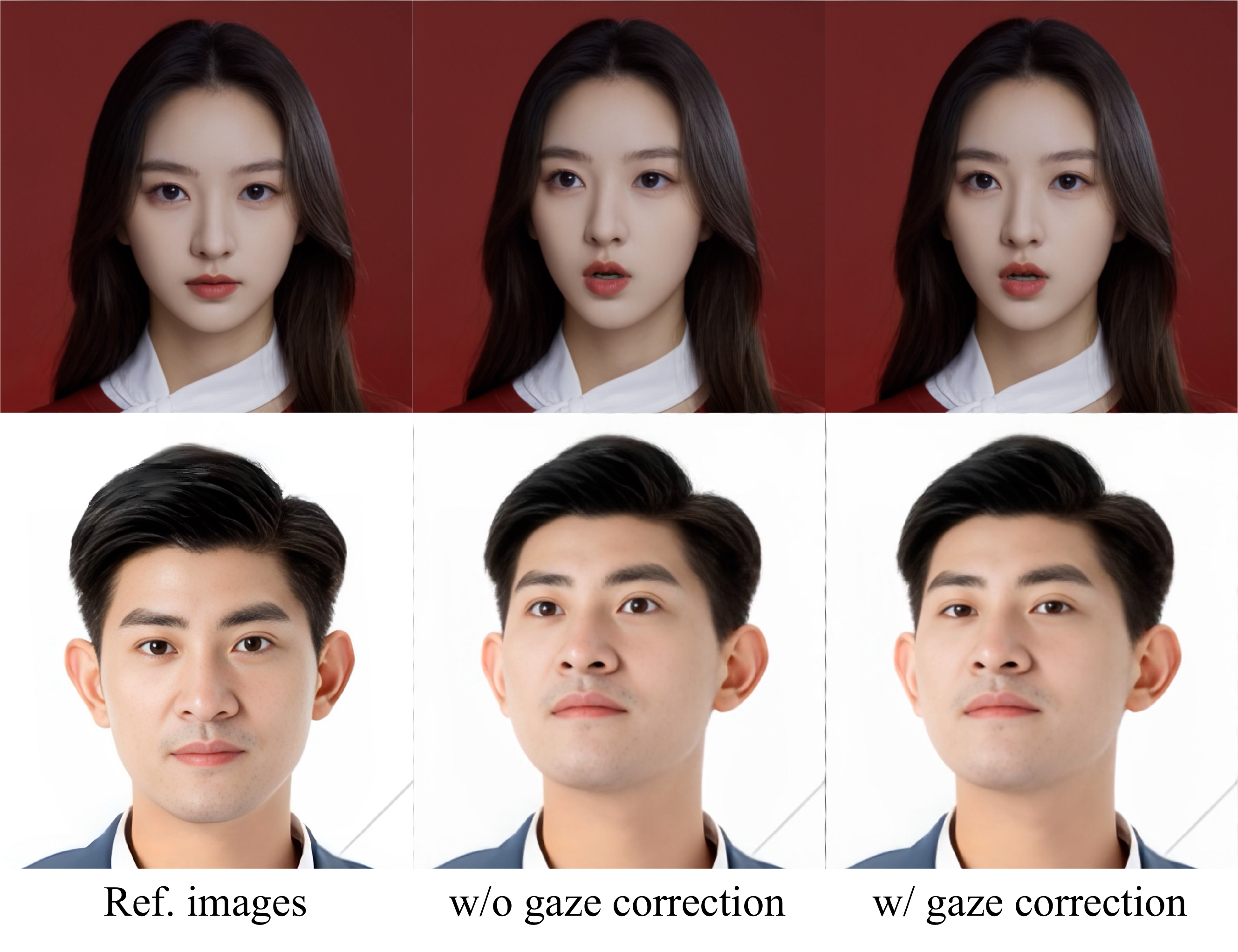}

  \caption{The gaze adjustment decouples gaze direction from the head pose, enabling the avatar's eyes to focus on the camera, producing more natural and engaging eye contact.}
  \label{fig:gaze}
\end{figure}

\subsection{Ablation Study}

\textbf{Conditional Information.}
We evaluate the impact of different controlling signals on the end-to-end generation results. As shown in Table~\ref{tab:abs}, when canonical keypoints are excluded~(\textbf{w/o C-kp}), all metrics drop, especially for FVD and Sync. This suggests that the identity-related geometry provided by canonical keypoints enhances the learning process of diffusion, particularly in adapting generated motion to the target identity. Conversely, imperfections in motion generation directly affect dynamic visual quality and lip synchronization accuracy.

When emotion labels are omitted~(\textbf{w/o Emo}), we observe a significant decline across all metrics. Without direct emotional control, learning facial expressions solely from audio becomes more challenging, resulting in a decrease in video quality and lip synchronization. In the supplementary materials, we demonstrate the consequent issues such as abrupt expression changes and exaggerated facial expressions, which also impact identity consistency throughout the video.

\textbf{Adaptive Loss Weights.}
Without adaptive loss weights~(\textbf{w/o Ada-w}), the problem of unbalanced convergence between pose and expression becomes severe, hindering the model to learn the correspondence between audio and facial motion. We observe that the generated videos exhibit minimal facial motions, resulting in poor Sync and FVD scores, while CSIM and FID metrics are less affected.

\textbf{Gaze Adjustment.}
As shown in Figure~\ref{fig:gaze}, without gaze adjustment, the gaze direction is rigidly bound to the head pose. The gaze adjustment decouples gaze direction from the head pose, enabling the avatar's eyes to focus on the camera, producing more natural and engaging eye contact.

\begin{table}
  \caption{The time consumption of single-step inference in different modules and corresponding real-time factor (RTF).}
  \label{tab:cost1}
  \centering
  \begin{tabular}{lccc}
    \toprule
    Module & Audio2Feat & Motion DiT & Face Rendering  \\
    \midrule
    inference & 23ms & 62ms & 15ms \\
    RTF & 0.115 & 0.310 & 0.375 \\
  \bottomrule
\end{tabular}
\end{table}

\begin{table}
  \caption{Inference performance for offline video output and online streaming output at the head region (512 $\times$ 512) and full-body scales (1920 $\times$ 1080).}
  \label{tab:cost2}
  \centering
  \begin{tabular}{lcc}
    \toprule
    Pipeline &  RTF  & FFD \\
    \midrule
    talking-head offline & 0.635 & - \\
    talking-head online & 0.895 & 385ms \\
    full-body offline & 0.648 & - \\
    full-body online & 0.914 & 392ms \\
  \bottomrule
\end{tabular}
\end{table}

\subsection{Inference Performance}

We separately evaluate the inference performance of the proposed method from both modular and end-to-end perspectives. We evaluate module-wise performance using both inference time and valid Real-Time Factor (RTF), which represents the single-step inference time divided by the valid segment length (excluding the overlapping part). The test environment is a 12-core Intel(R) Xeon(R) Platinum 8369B CPU @ 2.90GHz, 1 NVIDIA A100 GPU, and 100G memory.

Table~\ref{tab:cost1} shows that each module's RTF is under 1, guaranteeing realtime end-to-end inference for the pipeline. In Table~\ref{tab:cost2},  we present the inference performance for offline video output and online streaming output at two scales: the head region and full-body portrait. Compared to offline video output, online streaming output requires longer inter-segment overlap to ensure low First-Frame Delay (FFD), resulting in a higher RTF. Full-body output is only marginally slower than head-only output, primarily due to the process of integrating the head into the body. Our method ensures end-to-end inference with RTF below 1 and FFD under 400ms, meeting the requirements for realtime interactive scenarios.

\section{Conclusion}

We propose Ditto, a diffusion-based talking head framework that enables fine-grained controls and real-time inference. We devise a Conditional DiT and corresponding training strategy to address the issues in generating motion representations. We jointly optimize the holistic framework to enable streaming processing, real-time inference, and low first-frame delay. For the advancement of the community, we make our source code available for open-source use.

\normalsize
\bibliography{main}


\end{document}